\documentclass[accepted]{uai2023} %

\usepackage{microtype}
\usepackage{bmpsize}
\usepackage[hidelinks]{hyperref}
\usepackage{booktabs}
\hypersetup{
    colorlinks=true,
    linkcolor=blue,
    urlcolor=blue,
    citecolor=blue,
    anchorcolor=blue}

\usepackage{amsmath}
\usepackage{amssymb}
\usepackage{mathtools}
\usepackage{amsthm}
\usepackage{algorithmic,algorithm}

\usepackage[round]{natbib} %
\usepackage[inline]{enumitem}
\usepackage[table, dvipsnames]{xcolor}
\usepackage{color}

\usepackage{amsmath}
\usepackage{float}
\usepackage{adjustbox}
\usepackage{caption} 
\usepackage{mathtools, nccmath}
\usepackage{tikz}
\usepackage[algo2e, ruled,vlined,boxed,linesnumbered]{algorithm2e}
\SetArgSty{textnormal}
\usepackage{listings}
\usepackage{multicol}
\usepackage{wrapfig}
\usepackage{enumitem}
\usepackage{makecell}
\usepackage{upgreek}

\usepackage[toc,page]{appendix}      %

\usepackage{amsmath}		%
\usepackage{amssymb}		%
\usepackage{amsfonts}		%
\usepackage{amsthm}		%

\usepackage{mathtools}		%
\usepackage{dsfont}		%

\usepackage{acronym}		%

\newcommand{\acli}[1]{\textit{\acl{#1}}}		%
\newcommand{\acdef}[1]{\textit{\acl{#1}} \textup{(\acs{#1})}\acused{#1}}		%
\newcommand{\acdefp}[1]{\textit{\aclp{#1}} \textup{(\acsp{#1})}\acused{#1}}	%

\newcommand{\afterhead}{.}
\usepackage{titlesec}
\newcommand{\para}[1]{\vspace{-10pt}\smallskip\paragraph{\textbf{#1\afterhead}}}

\usepackage{quoting}			%
\quotingsetup{vskip=\medskipamount}

\usepackage{stmaryrd}		%
\usepackage{wasysym}		%

\usepackage{booktabs}		%

\usepackage[sort&compress,capitalize,nameinlink]{cleveref}		%
\crefname{assumption}{Assumption}{Assumptions}
\crefname{algo}{Algorithm}{Algorithms}
\crefname{example}{Example}{Examples}
\crefname{method}{Method}{Methods}
\creflabelformat{assumption}{\upshape(#2#1#3\upshape)}

\crefname{assumptionenum}{Assumption}{Assumptions}
\creflabelformat{assumptionenum}{#2#1#3}

\crefname{item}{}{}
\creflabelformat{item}{#2#1#3}

\crefname{eq}{}{}
\creflabelformat{eq}{\upshape(#2#1#3\upshape)}

\usepackage{thmtools}		%
\usepackage{thm-restate}		%

\newcommand{\debug}[1]{#1}		%

\definecolor{darkblue}{HTML}{1A254B}
\definecolor{lightblue}{HTML}{A7BED3}
\definecolor{blue}{HTML}{114083}
\definecolor{green}{HTML}{81B5AE}
\definecolor{pink}{HTML}{F2545B}
\definecolor{red}{HTML}{A4243B}
\definecolor{airforceblue}{rgb}{0.36, 0.54, 0.66}
\definecolor{thistle}{rgb}{0.85, 0.75, 0.85}
\definecolor{ticklemepink}{rgb}{0.99, 0.54, 0.67}
\definecolor{thulianpink}{rgb}{0.67, 0.24, 0.43}
\definecolor{tealblue}{rgb}{0.11, 0.36, 0.43}

\newcommand{\defeq}{\vcentcolon=}

\def\Ncal{\mathcal{N}}
\def\bP{\mathbf{P}}

\def\We{W_{\varepsilon}}

\DeclarePairedDelimiterX{\dotp}[2]{\langle}{\rangle}{#1, #2}

\DeclareMathOperator*{\argmin}{argmin}

\newcommand{\newmacro}[2]{\newcommand{#1}{\debug{#2}}}		%
\DeclarePairedDelimiter{\bracks}{[}{]}		%
\DeclarePairedDelimiter{\parens}{(}{)}		%

\DeclarePairedDelimiterX{\inner}[2]{\langle}{\rangle}{#1, #2}		%

\DeclarePairedDelimiter{\norm}{\lVert}{ \rVert}		%
\DeclarePairedDelimiterXPP{\twonorm}[1]{}{\lVert}{\rVert}{}{#1}		%
\DeclarePairedDelimiterXPP{\dnorm}[1]{}{\lVert}{\rVert}{_{\ast}}{#1}		%

\DeclarePairedDelimiterX{\braket}[2]{\langle}{\rangle}{#1,#2}		%

\DeclarePairedDelimiterX{\setdef}[2]{\{}{\}}{#1:#2}		%
\DeclarePairedDelimiterXPP{\exclude}[1]{\mathopen{}\setminus}{\{}{\}}{}{#1}

\newcommand{\alt}[1]{#1'}		%

\newcommand{\R}{\mathbb{R}}		%
\DeclareMathOperator{\dist}{dist}		%
\DeclareMathOperator{\one}{\mathds{1}}		%

\newmacro{\coef}{\lambda}		%
\newmacro{\dd}{\:\mathrm{d}}		%
\newmacro{\intR}{\int_{\R^{\vdim}}}		%
\newmacro{\intRR}{\int_{\R^{\vdim}  \times \R^{\vdim}  }}		%
\newmacro{\nn}{\nonumber}		%

\newcommand{\subs}{\leftarrow}      %

\newcommand{\eps}{\varepsilon}		%

\newmacro{\pexp}{p}		%
\newmacro{\qexp}{q}		%
\newmacro{\rexp}{r}		%

\newcommand{\ie}{i.e.,\xspace}		%

\newmacro{\mfd}{\mathcal{M}}		%
\newmacro{\curve}{\gamma}          %
\newmacro{\sect}{\mathcal{K}}    %

\newmacro{\sset}{\mathcal{S}}		%

\newmacro{\points}{\mfd}		%
\newmacro{\intpoints}{\points^{\circ}}		%
\newmacro{\point}{x}		%
\newmacro{\pointalt}{\alt\point}		%

\newmacro{\dpoints}{\mathcal{W}}		%
\newmacro{\dpoint}{w}		%
\newmacro{\dpointalt}{\alt\dpoint}		%

\newmacro{\base}{p}		%
\newmacro{\basealt}{q}		%

\newmacro{\open}{\mathcal{U}}		%
\newmacro{\closed}{\mathcal{C}}		%
\newmacro{\cpt}{\mathcal{K}}		%
\newmacro{\nbhd}{\mathcal{U}}		%

\newmacro{\start}{1}		%
\newmacro{\halfafterstart}{3/2}		%
\newmacro{\afterstart}{2}		%
\newmacro{\running}{\start,\afterstart,\dotsc}		%
\newmacro{\halfrunning}{\start,\halfafterstart,\dotsc}

\newmacro{\runalt}{k}		%
\newmacro{\run}{n}		%
\newmacro{\nRuns}{T}		%
\newmacro{\runs}{\mathcal{\nRuns}}		%

\newmacro{\state}{Z}		%
\newmacro{\dstate}{Y}		%

\newmacro{\vecspace}{\R^{\vdim}}		%

\newmacro{\coord}{i}		%
\newmacro{\vdim}{d}		%
\newmacro{\vvec}{v}		%
\newmacro{\bvec}{e}		%
\newmacro{\bvecs}{\mathcal{E}}		%

\newmacro{\subspace}{\mathcal{W}}		%
\newmacro{\wvec}{w}		%
\newmacro{\subdim}{m}		%

\newmacro{\tanhull}{\mathcal{Z}}		%
\newmacro{\tanvec}{z}		%

\newcommand{\dual}[1]{#1^{\ast}}		%
\newmacro{\dspace}{\dual\vecspace}		%
\newmacro{\dvec}{v}		%
\newmacro{\dbvec}{\eps}		%

\newmacro{\mat}{M}		%
\newmacro{\eye}{I}		%

\DeclareMathOperator{\ex}{\mathbb{E}}		%
\DeclareMathOperator{\prob}{\mathbb{P}}		%

\newmacro{\seed}{\omega}		%
\newmacro{\seeds}{\Omega}		%
\newmacro{\history}{\mathcal{H}}		%

\newmacro{\sample}{\omega}		%
\newmacro{\samples}{\Omega}		%
\newmacro{\filter}{\mathcal{F}}		%
\newmacro{\probspace}{(\samples,\filter,\prob)}		%

\newmacro{\event}{\mathcal{E}}       %
\newmacro{\eventalt}{\mathcal{H}}       %

\newmacro{\mean}{\mu}		%
\newmacro{\sdev}{\sigma}		%
\newmacro{\variance}{\sdev^{2}}		%

\newmacro{\dkl}{D_{\mathrm{KL}}}		%

\DeclarePairedDelimiterXPP{\exof}[1]{\ex}{[}{]}{}{%
 #1}

\DeclarePairedDelimiterXPP{\probof}[1]{\prob}{(}{)}{}{%
 #1}

\newmacro{\gmat}{g}		%
\newmacro{\gdist}{\dist_{\gmat}}
\newmacro{\ball}{\mathbb{B}}		%
\newmacro{\sphere}{\mathbb{S}}		%

\newmacro{\mbase}{\mu}
\newmacro{\m}{\mbase_0}     %
\newmacro{\malt}{\mbase_\horizon}     %

\newmacro{\covarbase}{\Sigma}
\newmacro{\covar}{\covarbase_0}     %
\newmacro{\covaralt}{\covarbase_\horizon}     %

\newmacro{\ctime}{t}
\newmacro{\ctimealt}{s}
\newmacro{\horizon}{1}

\newmacro{\ratiosym}{r}    %

\newmacro{\scalingbase}{\mathrm{QV}}     %

\newmacro{\KLbase}{D_{\mathrm{KL}}}
\newcommand{\KL}[2]{ \KLbase\parens*{ #1 \Vert #2 } }

\newmacro{\sdebase}{ X }
\newmacro{\tinv}{\tau}

\newmacro{\testfbase}{u}
\newmacro{\generator}{\mathcal{L}_{\ctime}}

\usepackage{xparse}
\NewDocumentCommand{\testf}{ O{\ctime} O{\point} }{ \testfbase\parens*{#1,#2} }

\newmacro{\dconst}{\lambda}     %
\newmacro{\sconst}{\mathbf{v}}     %
\newmacro{\vconst}{\omega}     %

\newmacro{\qvbase}{\mathrm{q}}

\newmacro{\subVPfbase}{ \beta }

\newmacro{\refsdebase}{ Y }     %
\newcommand{\refsde}[1][\ctime]{ \refsdebase_{ #1 } }

\newmacro{\wiescalebase}{ g }

\newmacro{\QVbase}{ \mathrm{qv} }

\newmacro{\driftbase}{  c  }

\newmacro{\shiftbase}{  \alpha  }

\newmacro{\volatbase}{  g  }
\newcommand{\volat}[1][\ctime]{ \volatbase_{ #1 }  }
\newcommand{\volatsq}[1][\ctimealt]{ \volatbase^2_{ #1 }  }

\newmacro{\refprobase}{\mathbb{Q}}     %
\newcommand{\refpro}[1][\ctime]{\refprobase_{ #1 }}
\newmacro{\refjoint}{\refprobase_{\mathrm{0\horizon}} }    %

\newmacro{\Wienerbase}{\mathbb{W}}     %
\newcommand{\dWiener}[1][\ctime]{ \dd \Wienerbase_{#1} }     %

\newmacro{\aggtimebase}{  \tau  }

\newcommand{\aggtimesqinv}[1][\ctimealt]{ \aggtimebase^{-2}_{ #1 }  }

\newmacro{\mrsdebase}{ \eta}
\NewDocumentCommand{\mYcinit}{ O{\ctime}  }{  \mrsdebase\parens*{#1 } } %

\newmacro{\kernelbase}{ \kappa}
\NewDocumentCommand{\kernel}{ O{\ctime} O{\ctime'}  }{  \kernelbase\parens*{#1, #2} }%

\newmacro{\intdasq}{ \int_0^\ctime {\aggtimesqinv[\ctimealt]}{\volatsq} \dd \ctimealt }
\newmacro{\intdasqT}{ \int_0^\horizon {\aggtimesqinv[\ctimealt]}{\volatsq} \dd \ctimealt }

\newmacro{\ssstyle}{\scriptscriptstyle}
\newmacro{\sssNcal}{\ssstyle\Ncal}

\newmacro{\solbase}{\star}

\newmacro{\Cstar}{C_{\sdev_{\solbase}}}

\newmacro{\pbase}{\mathbb{P}}     %
\newmacro{\Pinit}{\pbase_{{0}}}     %
\newmacro{\Pend}{\pbase_{{\horizon}}}     %
\newcommand{\Pmargin}[1][\ctime]{ \pbase_{{#1}} }     %
\newcommand{\Psol}[1][\ctime]{ \pbase^{\solbase}_{ #1 } }
\newmacro{\Pjoint}{ \pbase_{ \mathrm{0\horizon}} }

\newmacro{\distbase}{ \pbase }
\newmacro{\ini}{ {0} }
\newmacro{\distinit}{ \hat{\distbase}_{ \ini } }
\newmacro{\en}{ {\horizon} }
\newmacro{\distend}{ \hat{\distbase}_{ \en} }

\newmacro{\efftrbase}{\rho}    %

\newmacro{\paramf}{ \theta }
\newmacro{\SBfbase}{ Z }
\newmacro{\paramb}{ \phi }
\newmacro{\SBbbase}{ \hat{\SBfbase} }
\NewDocumentCommand{\SBf}{ O{\ctime} O{\point} O{\paramf} }{ \SBfbase_{#1}^{#3}\parens{#2} }
\NewDocumentCommand{\SBb}{ O{\ctime} O{\point} O{\paramb} }{ \SBbbase_{#1}^{#3}\parens{#2} }

\newmacro{\GSBfbase}{ f_{\sssNcal} } %
\newmacro{\GSBbbase}{ \hat{\GSBfbase}}%
\NewDocumentCommand{\GSBf}{ O{\ctime} O{\point} }{ \GSBfbase\parens*{#1,#2} }

\newmacro{\tshiftbase}{ \zeta }%

\newmacro{\dt}{ \dd \ctime}

\newmacro{\loss}{\ell}
\NewDocumentCommand{\lossf}{ O{\point_{\horizon}} O{\paramf} }{  \loss\parens*{ #1; #2 }}

\NewDocumentCommand{\lossb}{ O{\point_{0}} O{\paramb} }{  \loss\parens*{ #1; #2 }}

\newmacro{\caching}{M}
\newmacro{\outeriter}{K_\textup{out}}
\newmacro{\inneriter}{K_\textup{in}}

\newmacro{\pretriterf}{K_{\paramf}}
\newmacro{\pretriterb}{K_{\paramb}}

\newmacro{\lrbase}{\gamma}
\newmacro{\lrf}{\lrbase_{\paramf}}
\newmacro{\lrb}{\lrbase_{\paramb}}

\newmacro{\Ninit}{\Ncal_{\ini}}
\newmacro{\Nend}{\Ncal_{\en}}

\newmacro{\tdriftbase}{ f }

\NewDocumentCommand{\tdrift}{ O{\ctime} O{\refsde} }{ \tdriftbase\parens*{#1,#2} }

\newmacro{\SB}{ {\scriptscriptstyle\textup{SB}} } %

\newcommand{\x}{\mathbf{x}}
\newcommand{\Pssol}{\pi^\star}

\newacro{LHS}{left-hand side}
\newacro{RHS}{right-hand side}
\newacro{iid}[i.i.d.]{independent and identically distributed}
\newacro{lsc}[l.s.c.]{lower semi-continuous}

\newacro{GAN}{generative adversarial network}
\newacro{NN}{neural network}
\newacro{FTRL}{``follow the regularized leader''}
\newacro{wp1}[w.p.$1$]{with probability $1$}

\newacro{SDE}{stochastic differential equation}
\newacro{SB}{Schr\"odinger bridge}
\newacro{GSB}[GSB]{Gaussian Schr\"odinger bridge}

\newacro{SGM}{score-based generative model}

\newacro{SMLD}{score matching with Langevin dynamics}
\newacro{DDPM}{denoising diffusion probabilistic model}

\newacro{OU}{Ornstein\textendash Uhlenbeck}
\newacro{BM}{Brownian motion}
\newacro{BDT}{Black–Derman–Toy}

\newacro{VESDE}[VE SDE]{variance exploding \ac{SDE}}
\newacro{VPSDE}[VP SDE]{variance preserving \ac{SDE}}
\newacro{DSB}{diffusion Schr\"odinger bridge}
\newacro{IPF}{iterative proportional fitting}

\newmacro{\acroalg}{\textsc{GSBflow}}   %

\theoremstyle{plain}

\title{Aligned Diffusion Schr\"odinger Bridges}

\author[1,2]{Vignesh Ram Somnath$^*$}
\author[1,3]{Matteo Pariset$^*$}
\author[1]{Ya-Ping Hsieh}
\author[2]{\\Maria Rodriguez Martinez}
\author[1]{Andreas Krause}
\author[1]{Charlotte Bunne}
\affil[1]{%
    Department of Computer Science\\
    ETH Z\"urich
}
\affil[2]{%
    IBM Research Z\"urich
}
\affil[3]{%
    Department of Computer Science\\
    EPFL
  }
  
\begin{document}
\renewcommand\ttdefault{lmtt}
\maketitle

\begin{abstract}
\vspace{-15pt}
Diffusion Schr\"odinger bridges (\acsu{DSB}) have recently emerged as a powerful framework for recovering stochastic dynamics via their marginal observations at different time points. Despite numerous successful applications, existing algorithms for solving \acp{DSB} have so far failed to utilize the structure of \emph{aligned} data, which naturally arises in many biological phenomena. In this paper, we propose a novel algorithmic framework that, for the first time, solves \acp{DSB} while respecting the data alignment. Our approach hinges on a combination of two decades-old ideas: The classical Schr\"odinger bridge theory and Doob's \emph{$h$-transform}. Compared to prior methods, our approach leads to a simpler training procedure with lower variance, which we further augment with principled regularization schemes. This ultimately leads to sizeable improvements across experiments on synthetic and real data, including the tasks of predicting conformational changes in proteins and temporal evolution of cellular differentiation processes.
\end{abstract}

\vspace{-5pt}
\footnotetext[1]{Equal contribution.}
\section{Introduction}
\label{sec:intro}
\vspace{-5pt}
\looseness -1 The task of transforming a given distribution into another lies at the heart of many modern machine learning applications such as single-cell genomics \citep{tong2020trajectorynet, schiebinger2019optimal, bunne2022supervised}, meteorology \citep{fisher2009data}, and robotics \citep{chen2021optimal}. 
To this end, \aclp{DSB} \citep{de2021diffusion,chen2021likelihood,vargas2021solving,liu2022deep} have recently emerged as a powerful paradigm due to their ability to generalize prior deep diffusion-based models, notably \acl{SMLD}
\citep{song2019generative,song2020score} and \aclp{DDPM} \citep{ho2020denoising}, which have achieved the state-of-the-art on many generative modeling problems.

\begin{figure}[t]
    \centering
    \includegraphics[width=\linewidth]{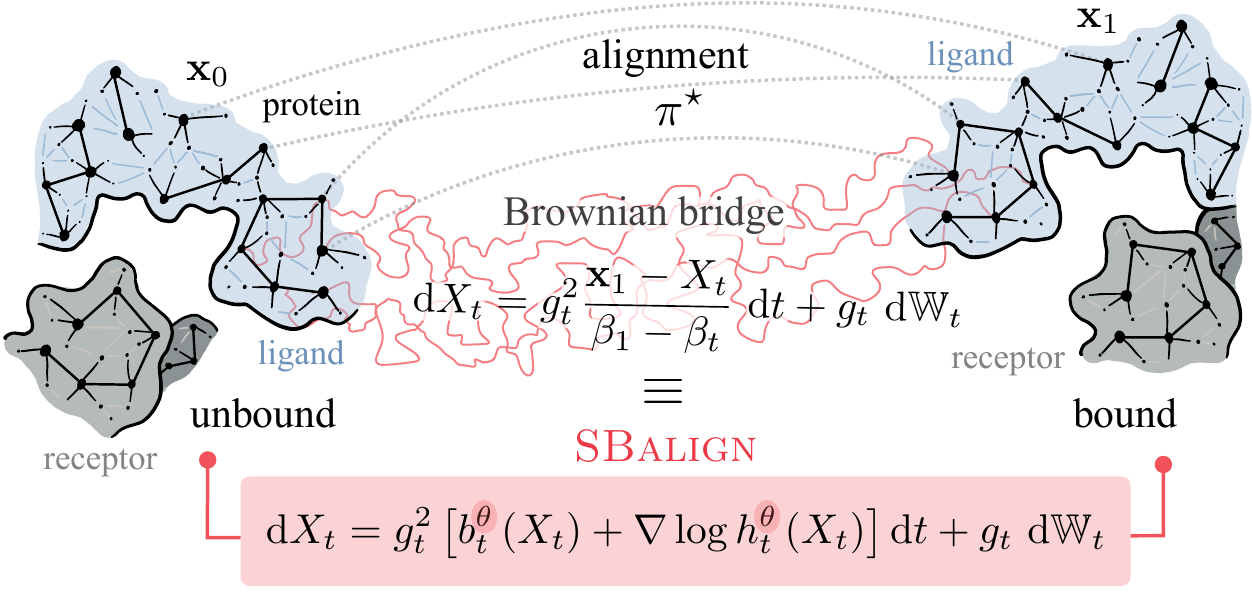}
    \caption{Overview of \textsc{SBalign}: In biological tasks such as protein docking, one is naturally provided with {\em aligned} data in the form of unbound and bound structures of participating proteins. Our goal is to therefore recover a stochastic trajectory from the unbound ($\x_0$) to the bound ($\x_1$) structure. To achieve this, we connect the characterization of an SDE conditioned on $\x_0$ and $\x_1$ (utilizing the Doob's \emph{$h$-transform}) with that of a Brownian bridge between $\x_0$ and $\x_1$ (classical Schr\"odinger bridge theory). We show that this leads to a simpler training procedure with lower variance and strong empirical results.}
    \label{fig:overview_proteins}
\end{figure}

\looseness -1 Despite the success of \acp{DSB} solvers, a significant limitation of existing frameworks is that they fail to capture the \emph{alignment} of data: If $\distinit, \distend$ are two (empirical) distributions between which we wish to interpolate, then a tacit assumption in the literature is that the dependence of $\distinit$ and $\distend$ is unknown and somehow has to be recovered. Such an assumption, however, ignores important scenarios where the data is \emph{aligned}, meaning that the samples from $\distinit$ and $\distend$ naturally come in pairs $(\x^i_0,\x^i_1)_i^{N}$, which is common in many biological phenomena. Proteins, for instance, undergo conformational changes upon interactions with other biomolecules (protein docking, see Fig.~\ref{fig:overview_proteins}). The goal is to model conformational changes by recovering a (stochastic) trajectory $\x_t$ based on the positions observed at two-time points $\left(\x_0, \x_1\right)$. Failing to incorporate this alignment would mean that we completely ignore information on the correspondence between the initial and final points of the molecules, resulting in a much harder problem than necessary.

Beyond, the recent use of SBs has been motivated by an important task in molecular biology: Cells change their molecular profile throughout developmental processes \citep{schiebinger2019optimal,bunne2021jkonet} or in response to perturbations such as cancer drugs \citep{lotfollahi2019scgen,bunne2021learning}. As most measurement technologies are destructive assays, i.e., the same cell cannot be observed twice nor fully profiled over time, these methods aim at reconstructing cell dynamics from \emph{unpaired} snapshots.
Recent developments in molecular biology, however, aim at overcoming this technological limitation. For example, \citet{chen2022live} propose a transcriptome profiling approach that preserves cell viability. \citet{weinreb2020lineage} capture cell differentiation processes by clonally connecting cells and their progenitors through barcodes (see illustrative Figure in Supplement).

\looseness -1 Motivated by these observations, the goal of this paper is to propose a novel algorithmic framework for solving \acp{DSB} with (partially) \emph{aligned} data. Our approach is in stark contrast to existing works which, due to the lack of data alignment, all rely on some variants of \acdef{IPF} \citep{fortet1940resolution, kullback1968probability} and are thus prone to numerical instability. On the other hand, via a combination of the original theory of Schr\"odinger bridges \citep{schrodinger1931umkehrung,leonard2013survey} and the key notion of Doob's \emph{$h$-transform} \citep{doob1984classical, rogers2000diffusions}, we design a novel loss function that completely bypasses the \ac{IPF} procedure and can be trained with much lower variance.

To summarize, we make the following contributions:
\begin{itemize}[topsep=0pt]
\item To our best knowledge, we consider, for the first time, the problem of interpolation with \emph{aligned} data. We rigorously formulate the problem in the \ac{DSB} framework.

\item Based on the theory of Schr\"odinger bridges and $h$-transform, we derive a new loss function that, unlike prior work on \acp{DSB}, does not require an \ac{IPF}-like procedure to train. We also propose principled regularization schemes to further stabilize training.

\item We describe how interpolating aligned data can provide better reference processes for use in classical \acp{DSB}, paving the way to hybrid aligned/non-aligned \acp{SB}.

\item \looseness -1 We evaluate our proposed framework on both synthetic and real data. For experiments utilizing real data, we consider two tasks where such aligned data is naturally available. The first is the task of developmental processes in single-cell biology, and the second involves protein docking. While diffusion models have been applied to model the relative orientation of proteins during docking, they have not been extended to model protein flexibility. We showcase a proof-of-concept application of our method on modelling conformational changes between unbound and bound states of a protein. Our method demonstrates a considerable improvement over prior methods across various metrics, thereby substantiating the importance of taking the data alignment into account.

\end{itemize}

\para{Related work}

Solving \acp{DSB} is a subject of significant interest in recent years and has flourished in a number of different algorithms \citep{de2021diffusion,chen2021likelihood,vargas2021solving,bunne2022recovering,liudeep2022}. However, all these previous approaches focus on  \emph{unaligned} data, and therefore the methodologies all rely on \ac{IPF} and are hence drastically different from ours. In the experiments, we will demonstrate the importance of taking the alignment of data into consideration by comparing our method to these baselines.

An important ingredient in our theory is Doob's $h$-transform, which has recently also been utilized by \citep{liu2023learning, heng2021simulating} to solve the problem of conditional diffusion. However, their fundamental motivation is different from ours. \citet{heng2021simulating} focus on learning the $h$-transform for a given drift function by approximating two score functions, while \citet{liu2023learning} focus on learning the drift of the diffusion model and the $h$-transform \emph{together}. In contrast, our goal is to read off the drift \emph{from} the $h$-transform with the help of {\em aligned data}, and use the learnt drift for unconditioned simulation.

To the best of our knowledge, the concurrent work of \citet{tong2023conditional} is the only existing framework that can tackle aligned data, which, however, is not their original motivation. In the context of solving \acp{DSB}, their algorithm can be seen as learning a vector field that generates the correct \emph{marginal} probability \citep[cf.][Proposition 4.3]{tong2023conditional}. Importantly, this is different from our aim of finding the \emph{pathwise} optimal solution of \acp{DSB}: If $(\x^i_{0,\textup{test}})_{i=1}^m$ is a test data set for which we wish to predict their destinations, then the framework of \citet{tong2023conditional} can only ensure that the marginal distribution $(\x^i_{1,\textup{test}})_{i=1}^m$ is correct, whereas ours is capable of predicting that $\x^i_{1,\textup{test}}$ is precisely the destination of $\x^i_{0,\textup{test}}$ for each $i$. This latter property is highly desirable in tasks like ML-accelerated protein docking.

To solve aligned \ac{SB} problems, we rely on mixtures of diffusion processes. Like in \citet{peluchetti2023diffusion}, we construct them from pairings and define an associated training objective inspired by score-based modeling. However, we represent the learned drift as a sum of the solution to an SB problem ($b$) and a pairing-related term ($\nabla \log h$). We parametrize the second part of the drift with neural networks, unlike \citet{schauer2017guided} which use an auxiliary (simpler) process.

\section{Background}
\label{sec:background}

\para{Problem formulation}

Suppose that we are given access to i.i.d. \emph{aligned} data $(\x_0^i,\x_1^i)_{i=1}^N$, where the marginal distribution of $\x^i_0$'s is $\distinit$ and of $\x_1^i$'s is $\distend$. Typically, we view $\distinit$ as the empirical marginal distribution of a stochastic process observed at time $t= 0$, and likewise $\distend$ the empirical marginal observed at $t=\horizon$. The goal is to reconstruct the stochastic process $\Pmargin$ based on $(\x_0^i,\x_1^i)_{i=1}^N$, \ie to \emph{interpolate} between $\distinit$ and $\distend$.

Such a task is ubiquitous in biological applications. For instance, understanding how proteins dock to other biomolecules is of significant interest in biology and has become a topic of intense study in recent years \citep{ganea2022independent, tsaban2022harnessing, corso2022diffdock}. In the protein docking task, $\x_0^i$ represents the 3D structures of the unbound biomolecules, while $\x_1^i$ represents the 3D structure of the bound complex. Reconstructing a stochastic process that diffuses $\x_0^i$'s to $\x_1^i$'s is tantamount to recovering the energy landscape governing the docking process.  Similarly, in molecular dynamics simulations, we have access to trajectories $\left(\x_t^i\right)_{t \in [0, 1]}$, where $\x_0^i$ and $\x_1^i$ represent the initial and final positions of the $i$-th molecule respectively. Any learning algorithm using these simulations should be able to respect the provided alignment. 

\para{Diffusion Schr\"odinger bridges}

To solve the interpolation problem, in \cref{sec:Methods}, we will invoke the framework of \acp{DSB}, which are designed to solve interpolation problems with \emph{unaligned} data. More specifically, given two marginals $\distinit$ and $\distend$, the \ac{DSB} framework proceeds by first choosing a reference process $\refpro$ using prior knowledge, for instance a simple Brownian motion, and then solve the entropy-minimization problem over all stochastic processes $\Pmargin$:
\begin{equation}
\label{eq:SB}
\tag{SB}
\min_{ \substack{ \Pinit = \distinit, \; \Pend = \distend} } \KL{\Pmargin}{\refpro}.
\end{equation}

Despite the fact that many methods exist for solving \eqref{eq:SB}  \citep{de2021diffusion,chen2021likelihood,vargas2021solving,bunne2022recovering}, none of these approaches are capable of incorporating \emph{alignment} of the data. This can be seen by inspecting the objective \eqref{eq:SB}, in which the coupling information $(\x_0^i,\x_1^i)$ is completely lost as only its individual marginals $\distinit,\distend$ play a role therein. Unfortunately, it is well-known that tackling the marginals separately necessitates a forward-backward learning process known as the \acli{IPF} (IPF) procedure \citep{fortet1940resolution,kullback1968probability}, which constitutes the primary reason of high variance training, thereby confronting \acp{DSB} with numerical and scalability issues. Our major contribution, detailed in the next section, is therefore to devise the first algorithmic framework that solves the interpolation problem with aligned data \emph{without} resorting to IPF.

\section{Aligned Diffusion Schr\"odinger Bridges}
\label{sec:Methods}
\newcommand{\fdrift}{b_t}
\newcommand{\doob}{h_t}
\newcommand{\doobs}{h_{t,\reg}}
\newcommand{\Loss}{L}
\newcommand{\reg}{\tau}
\newcommand{\cvolatbase}{\beta}
\newcommand{\cvolat}[1][\ctime]{\cvolatbase_{#1}}

In this section, we derive a novel loss function for \acp{DSB} with aligned data by combining two classical notions: The theory of \aclp{SB} \citep{schrodinger1931umkehrung,leonard2013survey,chen2021stochastic} and Doob's $h$-transform \citep{doob1984classical, rogers2000diffusions}. We then describe how solutions to DSBs with aligned data can be leveraged in the context of classical DSBs.

\subsection{Learning aligned diffusion Schr\"odinger bridges}
\para{Static SB and aligned data}

Our starting point is the simple and classical observation that \eqref{eq:SB} is the continuous-time analogue of the \emph{entropic optimal transport}, also known as the \emph{static} \acl{SB} problem \citep{leonard2013survey,chen2021stochastic,Peyre2019computational}:
\begin{equation}
\label{eq:static-SB}
\Pssol \defeq \argmin_{ \substack{ \Pinit = \distinit, \; \Pend = \distend} } \KL{\mathbb{P}_{0,1}}{\refprobase_{0,1}}
\end{equation}where the minimization is over all \emph{couplings} of $\distinit$ and $\distend$, and $\refprobase_{0,1}$ is simply the joint distribution of $\refpro$ at $t=0,\horizon$. In other words, if we denote by $\Psol$ the stochastic process that minimizes \eqref{eq:SB}, then the joint distribution $\Psol[0,\horizon]$ necessarily coincides with the $\Pssol$ in \eqref{eq:static-SB}. Moreover, since in \acp{DSB}, the data is always assumed to arise from $\Psol$, we see that:
\begin{quote}
The \emph{aligned} data $(\x_0^i,\x_1^i)_{i=1}^N$ constitutes samples of $\Pssol$.
\end{quote}
This simple but crucial observation lies at the heart of all derivations to come. 

Our central idea is to represent $\Psol$ via two different, but equivalent, characterizations, both of which involve $\Pssol$: That of a \emph{mixture} of reference processes with pinned end points, and that of conditional \acdefp{SDE}.

\para{$\Psol$ from $\Pssol$: $\refpro$ with pinned end points}

For illustration purposes, from now on, we will assume that the reference process $\refpro$ is a Brownian motion with diffusion coefficient $\volat$:\footnote{\looseness -1 Extension to more involved reference processes is conceptually straightforward but notationally clumsy. Furthermore, reference processes of the form \eqref{eq:gtWt} are dominant in practical applications \citep{song2020score, bunne2022recovering}, so we omit the general case. }
\begin{equation}
\label{eq:gtWt}
\dd \refpro = \volat \dWiener.
\end{equation}
In this case, it is well-known that $\refpro$ \emph{conditioned} to start at $\x_0$ and end at $\x_1$ can be written in another \ac{SDE} \citep{mansuy2008aspects, liu2023learning}:
\begin{equation}
\label{eq:BB}
\dd X_t = \volatsq[\ctime] \frac{\x_1-X_t}{\cvolat[\horizon]-\cvolat[\ctime]} \dt + \volat\dWiener
\end{equation}
where $X_0 = \x_0$ and %
\begin{equation}
\cvolat\defeq \int_0^\ctime \volatsq \dd s.
\end{equation}We call the processes in \eqref{eq:BB} the \emph{scaled Brownian bridges} as they generalize the classical Brownian bridge, which corresponds to the case of $\volat \equiv 1$.

The first characterization of $\Psol$ is then an immediate consequence the following classical result in \acl{SB} theory: Draw a sample $(\x_0, \x_1) \sim \Pssol$ and connect them via \eqref{eq:BB}. The resulting path is a sample from $\Psol$ \citep{leonard2013survey, chen2021stochastic}. In other words, $\Psol$ is a \emph{mixture} of scaled Brownian bridges, with the mixing weight given by $\Pssol$.

\para{$\Psol$ from $\Pssol$: \ac{SDE} representation}

Another characterization of $\Psol$ is that it is itself given by an \ac{SDE} of the form \citep{leonard2013survey, chen2021stochastic}
\begin{equation}
\label{eq:SB-SDE}
\dd X_t = \volatsq[\ctime]\fdrift(X_t) \dt + \volat\dWiener.
\end{equation}
Here, $\fdrift: \R^d \to \R^d$ is a time-dependent drift function that we wish to learn. 

Now, by Doob's h-transform, we know that the \ac{SDE} \eqref{eq:SB-SDE} \emph{conditioned} to start at $\x_0$ and end at $\x_1$ is given by another \ac{SDE}  \citep{doob1984classical,rogers2000diffusions}:
\begin{equation}
\label{eq:SB-SD-conditioned}
\dd X_t = \volatsq[\ctime]\bracks*{\fdrift(X_t) + \nabla \log \doob(X_t) }\dt +\volat \dWiener
\end{equation}
where $\doob(\x) \defeq \prob(X_1 = \x_1\vert X_t = \x)$ is the \emph{Doob's $h$ function}. Notice that we have suppressed the dependence of $\doob$ on $\x_0$ and $\x_1$ for notational simplicity.%

\para{Loss Function} Since both \eqref{eq:BB} and \eqref{eq:SB-SD-conditioned} represent $\Psol$, the solution of the \acp{DSB}, the two \acp{SDE} must coincide. To learn $\fdrift^\theta$, we thus consider the following loss:
\begin{equation}
\label{eq:loss_modified}
\Loss(\theta) \defeq \exof*{\int_0^1 \norm*{\frac{\x_1-X_t}{\cvolat[\horizon]-\cvolat[\ctime]}- \left(\fdrift^\theta + \nabla \log \doob^\theta(X_t)\right)}^2 \dt  }
\end{equation} for which $\fdrift$ is a minimizer. Note that, $\doob^\theta$ depends on $\fdrift^\theta$ as well as the drawn samples $(\x_0,\x_1)$. This is the case since $\doob$ is defined as an expectation using trajectories sampled under $\fdrift^\theta$ with given endpoints. Notice that \eqref{eq:loss_modified} bears a similar form as the popular score-matching objective employed in previous works \citep{song2019generative,song2020score}:
\begin{equation}
\label{eq:score_matching}
\Loss(\theta) \defeq \exof*{\int_0^1 \norm*{\nabla \log p(\x_t | \x_0)- s^\theta(X_t, t)}^2 \dt  },
\end{equation}
where the term $\frac{\x_1-X_t}{\cvolat[\horizon]-\cvolat[\ctime]}$ is akin to $\nabla \log p(\x_t | \x_0)$, while $\left(\fdrift^\theta + \nabla \log \doob^\theta(X_t)\right)$ corresponds to $s^\theta(X_t, t)$.

 \begin{algorithm}[t]
   \caption{\textsc{SBalign}}
   \label{alg:SBalign}
\begin{algorithmic}
   \STATE {\bfseries Input:} Aligned data $(\x^i_0,\x^i_1)_{i=1}^N$, learning rates $\lrf,\lrb$, number of iterations $K$ %
\smallskip
   \STATE Initialize $\paramf \subs \paramf_0$, $\paramb \subs \paramb_0$.
   \FOR{$k=1$ {\bfseries to} $K$} 
   \STATE Draw a mini-batch of samples from $(\x^i_0,\x^i_1)_{i=1}^N$
   \STATE Compute empirical average of \eqref{eq:loss_final} with mini-batch.
   \STATE Update $\paramb \subs \paramb - \lrb\nabla \Loss(\theta,\phi)$
   \STATE Update $\paramf \subs \paramf - \lrf\nabla \Loss(\theta,\phi)$
   \ENDFOR
\end{algorithmic}
\end{algorithm}

\para{Computing $\doob^\theta$}%

Inspecting $\doob$ in \eqref{eq:SB-SD-conditioned}, we see that, given $(\x_0,\x_1)$, it can be written as the conditional expectation of an indicator function:
\begin{equation}
\label{eq:h-semigroup}
\doob(\x) = \prob(X_1 = \x_1\vert X_t = \x) = \exof*{\one_{\{\x_1\}}\vert X_t = \x}
\end{equation}where the expectation is over \eqref{eq:SB-SDE}. Functions of the form \eqref{eq:h-semigroup} lend itself well to computation since it solves simulating the \emph{unconditioned} paths.
~Furthermore, in order to avoid overfitting on the given samples, it is customary to replace the ``hard'' constraint $\one_{\{\x_1\}}$ by its \emph{smoothed} version \citep{zhang2021path, holdijk2022path}: 
\begin{equation}
\label{eq:softdoob}
\doobs(\x) \defeq \exof*{\exp\parens*{-\frac{1}{2\reg}\norm{X_\horizon-\x_1}^2 }  \vert X_t = \x}.
\end{equation}Here, $\reg$ is a regularization parameter that controls how much we ``soften'' the constraint, and we have $\lim_{\reg\to 0} \doobs = \doob$.

Although the computation of \eqref{eq:softdoob} can be done via a standard application of the Feynman–Kac formula \citep{rogers2000diffusions}, an altogether easier approach is to parametrize $\doobs$ by a second neural network $m^{\phi}$ and perform alternating minimization steps on $\fdrift^\theta$ and $m^{\phi}$. This choice reduces the variance in training, since it avoids the sampling of unconditional paths described by \eqref{eq:SB-SDE} (see §\ref{app:sec:var_reduction} for a detailed explanation).

\begin{figure*}[t]
    \centering
    \includegraphics[width=\textwidth]{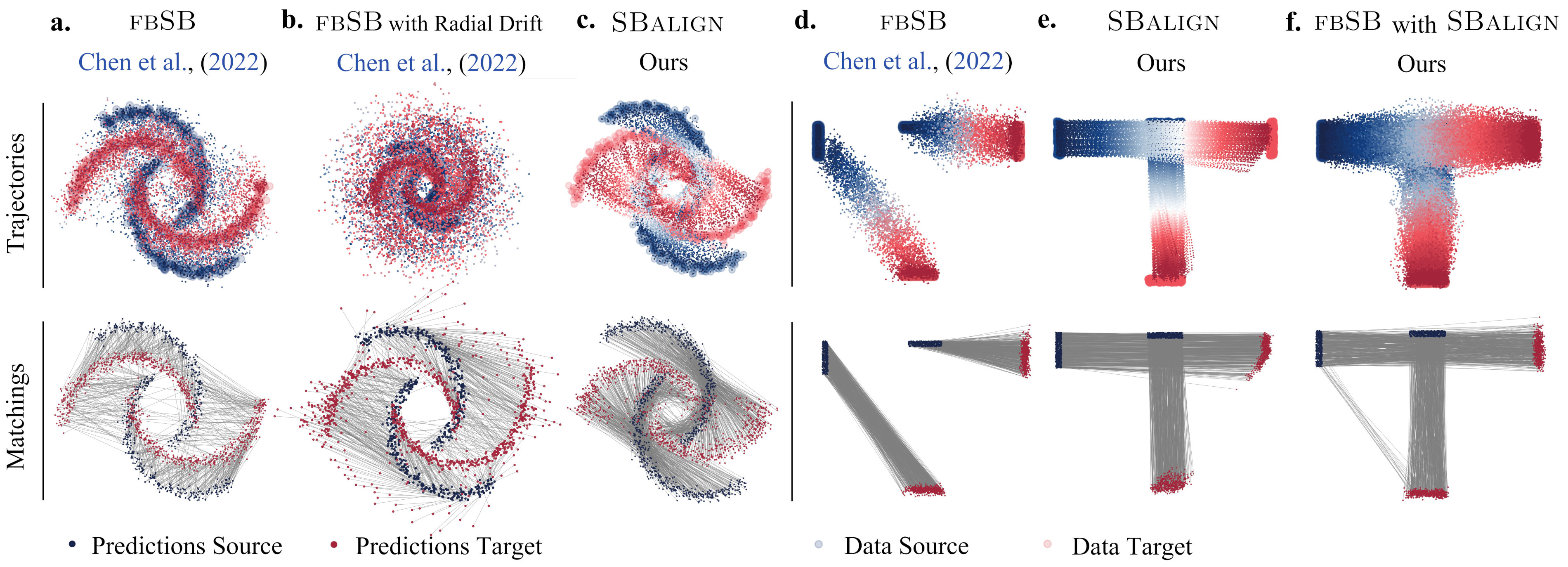}
    \caption{Experimental results on the Moon dataset (\textbf{a-c}) and T-dataset (\textbf{d-f}). The top row shows the trajectory sampled using the learned drift, and the bottom row shows the matching based on the learnt drift. Compared to other baselines, \textsc{SBalign} is able to learn an appropriate drift respecting the true alignment. (\textbf{f}) further showcases the utility of \textsc{SBalign}'s learnt drift as a suitable reference process to improve other training methods.}
    \label{fig:results_spiral}
\end{figure*}

\para{Regularization}
Since it is well-known that $\nabla \log\doob$ typically explodes when $\ctime\to 1$ \citep{liu2023learning}, it is important to regularize the behavior of $m^{\phi}$ for numerical stability, especially when $\ctime\to 1$. Moreover, in practice, it is desirable to learn a drift $\fdrift^\theta$ that respects the data alignment \emph{in expectation}: If $(\x_0,\x_1)$ is an input pair, then multiple runs of the \ac{SDE} \eqref{eq:SB-SDE} starting from $\x_0$ should, on average, produce samples that are in the proximity of $\x_1$. This observation implies that we should search for drifts whose corresponding $h$-transforms are diminishing.

A simple way to simultaneously achieve the above two requirements is to add an $\ell^2$-regularization term, resulting in the loss function:
\begin{align}
\label{eq:loss_final}
\Loss(\theta,\phi) &\defeq \mathbb{E} \Bigg[\int_0^1 \norm*{\frac{\x_1-X_t}{\cvolat[\horizon]-\cvolat[\ctime]}- \left(\fdrift^\theta + m^{\phi}(X_t)\right)}^2
\\ &\hspace{35mm}+ \lambda_t \norm{m^{\phi}(\x_t)}^2 \dt \Bigg]
\nonumber
\end{align}where $\lambda_t$ can either be constant or vary with time. The overall algorithm is depicted in \cref{alg:SBalign}.

\vspace{-5pt}

\subsection{Paired Schr\"odinger Bridges as prior processes}
\label{subsec:prior_drift}
\vspace{-8pt}

Our algorithm finds solutions to SBs on aligned data by relying on samples drawn from the (optimal) coupling $\pi^\star$. This is what differentiates it from classical SBs --which instead only consider samples from $\hat{\mathbb{P}}_0$ and $\hat{\mathbb{P}}_1$-- and plays a critical role in avoiding IPF-like iterates. However, \textsc{SBalign} reliance on samples from $\pi^\star$ may become a limitation, when the available information on alignments is insufficient. 

If the number of pairings is limited,  it is unrealistic to hope for an accurate solution to the aligned SB problem. However, the interpolation between $\hat{\mathbb{P}}_0$ and $\hat{\mathbb{P}}_1$ learned by \textsc{SBalign} can potentially be leveraged as a starting point to obtain a better reference process, which can then be used when solving a classical SB on the same marginals. In other words, the drift $b^\text{aligned}_t(X_t)$ learned through \textsc{SBalign} can be used \textit{as is} to construct a data-informed alternative $\tilde{\mathbb{Q}}$ to the standard Brownian motion, defined by paths:
\[
    \tilde{X}_t = b^\text{aligned}_t(\tilde{X}_t) dt + g_t dW_t
\]
Intuitively, solving a standard SB problem with $\tilde{\mathbb{Q}}$ as reference is beneficial because the (imperfect) coupling of marginals learned by \textsc{SBalign} ($\tilde{\mathbb{Q}}_{01}$) is, in general, closer to the truth than $\mathbb{Q}_{01}$.

Improving reference processes through pre-training or data-dependent initialization has been previously considered in the literature. For instance, both \citet{de2021diffusion} and \citet{chen2021likelihood} use a pre-trained reference process for challenging image interpolation tasks. This approach, however, relies on DSBs trained using the classical score-based generative modeling objective between a Gaussian and the data distribution. It, therefore, pre-trains the reference process on a related --but different-- process, i.e., the one mapping Gaussian noise to data rather than $\hat{\mathbb{P}}_0$ to $\hat{\mathbb{P}}_1$.
An alternative, proposed by \citet{bunne2022recovering} draws on the closed-form solution of SBs between two Gaussian distributions, which are chosen to approximate $\hat{\mathbb{P}}_0$ and $\hat{\mathbb{P}}_1$, respectively.
Unlike our method, these alternatives construct prior drifts by falling back to simpler and related tasks, or approximations of the original problem. We instead propose to shape a coarse-grained description of the drift based on alignments sampled directly from $\pi^\star_{01}$.

\section{Experiments}
\label{sec:experiments}

In this section, we evaluate \textsc{SBalign} in different settings involving 2-dimensional synthetic datasets, the task of reconstructing cellular differentiation processes, as well as predicting the conformation of a protein structure and its ligand formalized as rigid protein docking problem.

\subsection{Synthetic Experiments}
\label{sec:synthetic}

\begin{figure*}
    \centering
    \includegraphics[width=\textwidth]{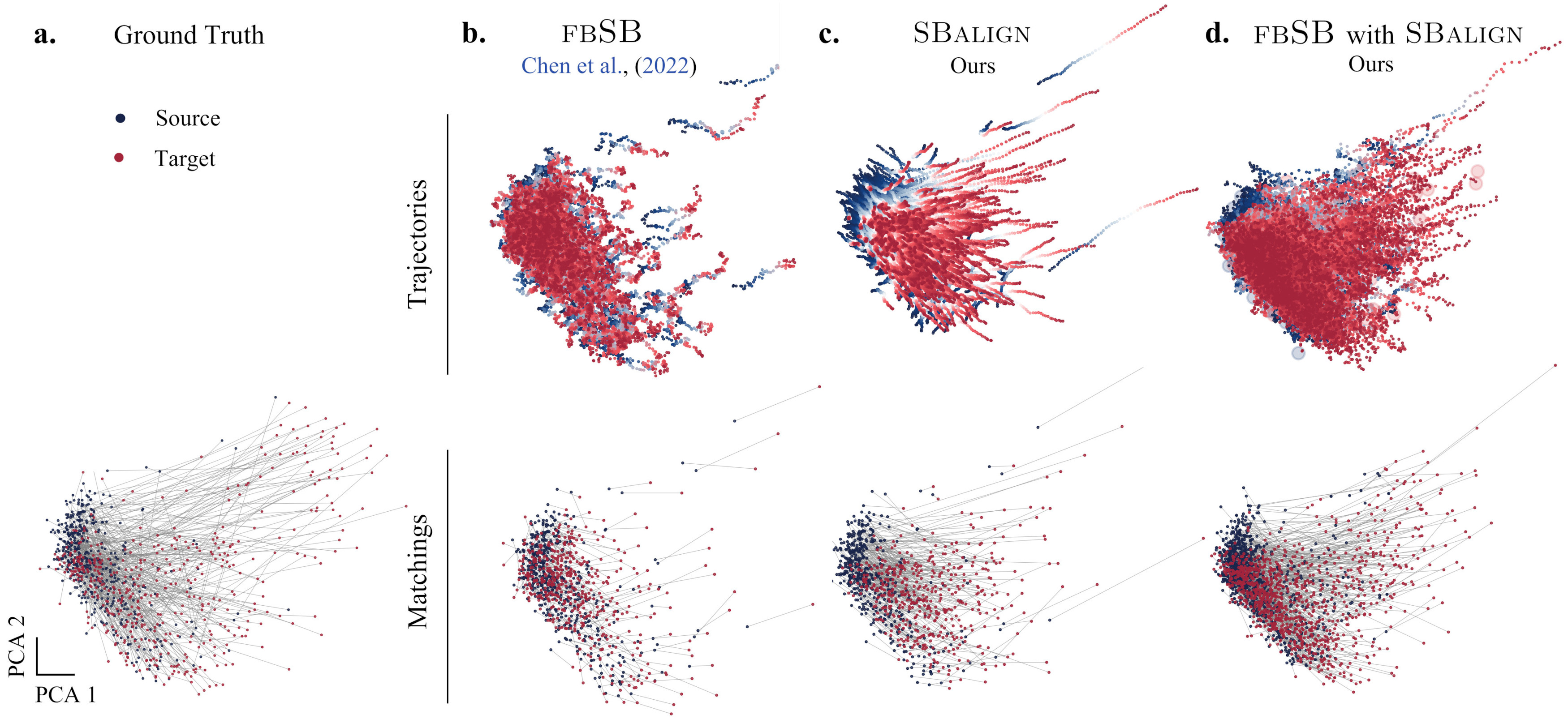}
    \caption{Cell differentiation trajectories based on (\textbf{a}) the ground truth and (\textbf{b-d}) learned drifts. \textsc{SBalign} is able to learn an appropriate drift underlying the true differentiation process while respecting the alignment. (\textbf{d}) Using the learned drift from \textsc{SBalign} as a reference process helps improve the drift learned by other training methods.}
    \label{fig:results_cell_traj}
\end{figure*}

We run our algorithm on two synthetic datasets (Figures in \S~\ref{app:datasets}), and compare the results with classic diffusion Schr\"odinger bridge models, i.e., the forward-backward SB formulation proposed by \cite{chen2021likelihood}, herein referred to as \textsc{fbSB}. We equip the baseline with prior knowledge, as elaborated below, to further challenge \textsc{SBalign}.

\para{Moon dataset}
The first synthetic dataset (Fig.~\ref{fig:results_spiral}a-c) consists of two distributions, each supported on two semi-circles ($\distinit$ drawn in \textit{blue} and $\distend$ in \textit{red}).
$\distend$ was obtained from $\distinit$ by applying a clockwise rotation around the center, i.e., by making points in the upper blue arm correspond to those in the right red one.
This transformation is clearly not the most likely one under the assumption of Brownian motion of particles and should therefore not be found as the solution of a classical SB problem. 
This is confirmed by \textsc{fbSB} trajectories (Fig.~\ref{fig:results_spiral}a), which tend to map points to their closest neighbor in $\distend$ (e.g., some points in the upper arm of $\distinit$ are brought towards the left rather than towards the right). 
While being a minimizer of \eqref{eq:SB}, such a solution completely disregards our prior knowledge on the alignment of particles, which is instead reliably reproduced by the dynamics learned by \textsc{SBalign} (Fig.~\ref{fig:results_spiral}c).

One way of encoding this additional information on the nature of the process is to modify $\refpro$ by introducing a clockwise radial drift, which describes the prior tangential velocity of particles moving circularly around the center.
Solving the classical SB with this updated reference process indeed generates trajectories that respect most alignments (Fig.~\ref{fig:results_spiral}b), but requires a hand-crafted expression of the drift that is only possible in very simple cases.

\para{T dataset}
In most real-world applications, it is very difficult to define an appropriate reference process $\refpro$, which respects the known alignment without excessively distorting the trajectories from a solution to \eqref{eq:SB}. This is already visible in simple examples like (Fig.~\ref{fig:results_spiral}d-f), in which the value of good candidate prior drifts at a specific location needs to vary wildly in time.
In this dataset, $\distinit$ and $\distend$ are both bi-modal distributions, each supported on two of the four extremes of an imaginary T-shaped area.
We target alignments that connect the two arms of the T as well as the top cloud with the bottom one. We succeed in learning them with \textsc{SBalign} (Fig.~\ref{fig:results_spiral}e) but unsurprisingly fail when using the baseline \textsc{fbSB} (Fig.~\ref{fig:results_spiral}d) with a Brownian motion prior.

\looseness -1 In this case, however, attempts at designing a better reference drift for \textsc{fbSB} must take into account the additional constraint that the horizontal and vertical particle trajectories intersect (see Fig.~\ref{fig:results_spiral}e), i.e., they cross the same area at times $t_h$ and $t_v$ (with $t_h > t_v$). This implies that the drift $b_t$, which initially points downwards (when $t < t_v$), should swiftly turn rightwards (for $t > t_h$).
Setting imprecise values for one of $t_h$ and $t_v$ when defining custom reference drifts for classical SBs would hence not lead to the desired result and, worse, would actively disturb the flow of the other particle group.

\looseness -1 As described in \S~\ref{subsec:prior_drift}, in presence of hard-to-capture requirements on the reference drift, the use of \textsc{SBalign} offers a remarkably easy and efficient way of learning a parameterization of it. For instance, when using the drift obtained by \textsc{SBalign} as reference drift for the computation of the SB baseline (\textsc{fbSB}), we find the desired alignments (Fig.~\ref{fig:results_spiral}f).

\begin{figure*}[t]
    \centering
    \includegraphics[width=\textwidth]{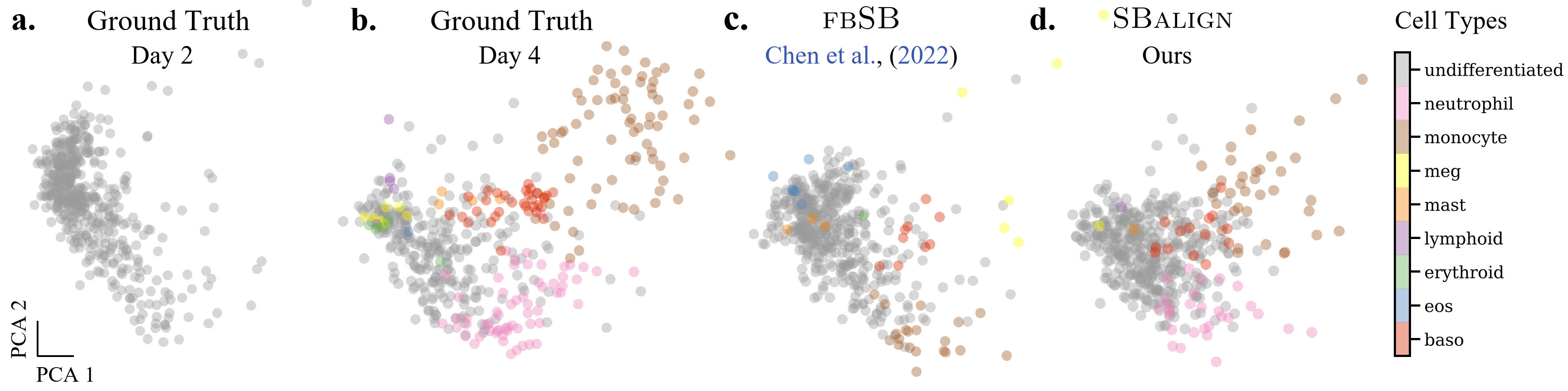}
    \caption{Cell type prediction on the differentiation dataset. All distributions are plotted on the first two principal components. \textbf{a-b:} Ground truth cell types on day 2 and day 4 respectively. \textbf{c-d:} \textsc{fbSB} and \textsc{SBalign} cell type predictions on day 4. \textsc{SBalign} is able to better model the underlying differentiation processes and capture the diversity in cell types.}
    \label{fig:results_cell_class}
\end{figure*}

\vspace{-5pt}
\subsection{Cell Differentiation}
\label{sec:cell}
\vspace{-5pt}

\looseness -1 Biological processes are determined through heterogeneous responses of single cells to external stimuli, i.e., developmental factors or drugs. Understanding and predicting the dynamics of single cells subject to a stimulus is thus crucial to enhance our understanding of health and disease and the focus of this task.
Most single-cell high-throughput technologies are destructive assays ---i.e., they destroy cells upon measurement--- allowing us to only measure \textit{unaligned} snapshots of the evolving cell population. Recent methods address this limitation by proposing (lower-throughput) technologies that keep cells alive after transcriptome profiling \citep{chen2022live} or that genetically tag cells to obtain a clonal trace upon cell division \citep{weinreb2020lineage}.

\para{Dataset} To showcase \textsc{SBalign}'s ability to make use of such (partial) alignments when inferring cell differentiation processes, we take advantage of the genetic barcoding system developed by \citet{weinreb2020lineage}. With a focus on fate determination in hematopoiesis, \citet{weinreb2020lineage} use expressed DNA barcodes to clonally trace single-cell transcriptomes over time. The dataset consists of two snapshots: the first, recorded on day 2, when most cells are still undifferentiated (see Fig.~\ref{fig:results_cell_class}a), and a second, on day 4, comprising many different mature cell types (see Fig.~\ref{fig:results_cell_class}b). Using \textsc{SBalign} as well as the baseline \textsc{fsSB}, we attempt to reconstruct cell evolution between day 2 and day 4, all while capturing the heterogeneity of emerging cell types. For details on the dataset, see \S~\ref{app:datasets}.

\begin{table}
    \caption{\textbf{Cell differentiation prediction results.} Means and standard deviations (in parentheses) of distributional metrics (maximum-mean-discrepancy (MMD), $\text{W}_{\epsilon}$), alignment-based metrics ($\ell_2$, RMSD), and cell type classification accuracy.  \vspace{-5pt}}
    \label{tab:results_cells}
     \centering
    \adjustbox{max width=\linewidth}{%
    \begin{tabular}{lccccc}
    \toprule
     & \multicolumn{5}{c}{\textbf{Cell Differentiation}} \\
    \cmidrule(lr){2-6}
    \textbf{Methods} & MMD $\downarrow$ & $\text{W}_\varepsilon \downarrow$ & $\ell_2(\text{PS}) \downarrow$ & RMSD $\downarrow$ & Class. Acc. $\uparrow$ \\
    \midrule
     \textsc{fbSB}& \makecell{1.55e-2\\(0.03e-2)} & \makecell{12.50\\(0.04)} & \makecell{4.08\\(0.04)} & \makecell{9.64e-1\\(0.02e-1)} & \makecell{56.2\%\\(0.7\%)} \\
     \makecell{\textsc{fbSB} with\\\textsc{SBalign}} & \makecell{5.31e-3\\(0.25e-3)} & \makecell{10.54\\(0.08)} & \makecell{0.99\\(0.12)} & \makecell{9.85e-1\\(0.07e-1)} & \makecell{47.0\%\\(1.5\%)} \\
     \textsc{\bf{\textsc{SBalign}}}& \makecell{1.07e-2\\(0.01e-2)} & \makecell{11.11\\(0.02)} & \makecell{1.24\\(0.02)} & \makecell{9.21e-1\\(0.01e-1)} & \makecell{56.3\%\\(0.7\%)} \\
     \bottomrule \vspace{-15pt}
    \end{tabular}
}
\end{table}

\para{Baselines} \looseness -1 We benchmark \textsc{SBalign} against previous \acp{DSB} such as \citep[\textsc{fbSB}]{chen2021likelihood}. Beyond, we compare \textsc{SBalign} in the setting of learning a prior reference process. Naturally, cell division processes and subsequently the propagation of the barcodes are very noisy. While this genetic annotation provides some form of assignment, it does not capture the full developmental process. We thus test \textsc{SBalign} in a setting where it learns a prior from such partial alignments and, plugged into \textsc{fbSB}, is fine-tuned on the full dataset.
 
\para{Evaluation metrics} To assess the performance of \textsc{SBalign} and the baselines, we monitor several metrics, which include distributional distances, i.e., MMD~\citep{gretton2012kernel} and $\text{W}_{\epsilon}$~\citep{cuturi2013sinkhorn}, as well as average (perturbation scores), i.e., $\ell_2(\text{PS})$ \citep{bunne2022supervised} and RMSD. Moreover, we also train a simple neural network-based classifier to annotate the cell type on day 4 and we report the accuracy of the predicted vs. actual cell type for all the models. See \S~\ref{app:metrics} for further details.

\looseness -1 \para{Results} \textsc{SBalign} finds matching between cell states on days 2 and 4 (Fig.~\ref{fig:results_cell_traj}c, bottom) which resemble the observed ones (Fig.~\ref{fig:results_cell_traj}a) but also reconstructs the entire evolution path of transcriptomic profiles (Fig.~\ref{fig:results_cell_traj}c, top).
It outperforms the baseline \textsc{fbSB} (Table \ref{tab:results_cells}) in all metrics: Remarkably, our method exceeds the performances of the baseline also on distributional metrics and not uniquely on alignment-based ones.
We also leverage \textsc{SBalign} predictions to recover the type of cells at the end of the differentiation process (Fig.~\ref{fig:results_cell_class}d). We do that by training a classifier on differentiated cells observed on day 4, and subsequently classify our predictions. %
While capturing the overall differentiation trend, \textsc{SBalign} (as well as \textsc{fbSB}) struggles to isolate rare cell types.
Lastly, we employ \textsc{SBalign} to learn a prior process from noisy alignments based on genetic barcode annotations. When using this reference process within \textsc{fbSB}, we learn an SB which compensates for inaccuracies stemming from the stochastic nature of cell division and barcode redistribution and which achieves better scores on distributional metrics (see Tab.~\ref{tab:results_cells}).
Further results can be found in \S~\ref{app:more_results}.

\vspace{-5pt}
\subsection{Conformational Changes in Protein Docking}
\vspace{-5pt}

Proteins are dynamic, flexible biomolecules that form complexes upon interaction with other biomolecules. 
The formation of complexes is guided by appropriate energetics, best orienting the participating proteins relative to each other, along with a dynamic alteration in structure (conformational changes) to form a stable complex. 

In ({\em computational}) protein docking, the goal is to predict the 3D structure of the bound (docked) state of a protein pair, given the unbound states of the corresponding proteins. These proteins are denoted (arbitrarily) as the ligand and receptor respectively.

While diffusion models have been applied for learning the appropriate orientations during docking \citep{corso2022diffdock, ketata2023diffdock}, modeling protein flexibility during docking is still an elusive problem. While it is possible to frame this problem as a ({\em conditional}) point cloud translation, an approach using DSBs is more natural since it leverages the dynamic and flexible nature of proteins and accounts for the underlying stochasticity in the process.

\vspace{5pt}
\para{Dataset} The task of modeling conformational changes starting from a given protein structure is largely unexplored, mainly due to the lack of high-quality large datasets. Here we utilize the recently proposed D3PM dataset \citep{peng2022d3pm} that provides protein structures before ({\em apo}) and after ({\em holo}) binding, covering various types of protein motions. The dataset was generated by filtering examples from the Protein Data Bank (PDB) corresponding to the same protein but bound to different biomolecules, with additional quality control criteria. For the scope of this work, we only focus on protein pairs where the provided Root Mean Square Deviation (RMSD) of the C$\alpha$ carbon atoms between unbound and bound 3D structures is $>3.0$\r{A}, which amounts to 2370 examples in the D3PM dataset.

For each pair of structures, we first identify common residues, and compute the RMSD between C$\alpha$ carbon atoms of the common residues after superimposing them using the Kabsch \citep{kabsch1976solution} algorithm, and only accept the structure if the computed C$\alpha$ RMSD is within a certain margin of the provided C$\alpha$ RMSD. The rationale behind this step is to only retain examples where we can reconstruct the RMSD values provided with the dataset. The above preprocessing steps give us a dataset with 1591 examples, which is then divided into a train/valid/test split of 1291/150/150 examples respectively. More details in \S~\ref{app:datasets-protein_conf}.  

\vspace{5pt}
\para{Baselines} Since the goal of the task is to predict 3D structures, our model must satisfy the relevant SE(3) symmetries of rotation and translations. To this end, we evaluate \textsc{SBAlign} against the \textsc{EGNN} model \citep{satorras2021n}, which satisfies the SE(3) symmetries and is a popular architecture used in many point-cloud transformation tasks \citep{satorras2021n, hoogeboom2022equivariant}. In particular, we consider the variant of the \textsc{EGNN} model proposed in \citet{xu2022geodiff}, owing to its strong performance on the molecule conformer generation task.

\vspace{5pt}
\para{Results} 
To evaluate our model, we report (Table~\ref{tab:results_conf_changes}) summary statistics of the RMSD between the C$\alpha$ carbon atoms of the predicted structure and the ground truth, and the fraction of predictions with RMSD values $<2.0, 5.0$ and $10.0$\r{A}.
\textsc{SBAlign} outperforms \textsc{EGNN} by a large margin and is able to predict almost 70\% examples with an RMSD$<5$\r{A}. One of the drawbacks attributed to diffusion models is their slow sampling speed, owing to multiple function calls to a neural network. Remarkably, our model is able to achieve impressive performance with just 10 steps of simulation. We leave it to future work to explore the tradeoff between sampling speed and quality of the predicted conformations.

\begin{table}
    \caption{\textbf{Conformational changes results.} RMSD between predicted and true structures in the bound state. The first term in the parentheses refers to the number of poses sampled, and the second term refers to the simulation steps for the trajectory.
    }
    \label{tab:results_conf_changes}
    \centering
    \adjustbox{max width=\linewidth}{%
    \begin{tabular}{lccc|ccc}
    \toprule
     & \multicolumn{6}{c}{\textbf{D3PM Test Set}} \\
     & \multicolumn{3}{c}{RMSD (\r{A})} & \multicolumn{3}{c}{\% RMSD (\r{A}) $< \tau$}  \\
    \cmidrule(lr){2-7}
    \textbf{Methods} & Median & Mean & Std & $\tau = 2$ & $\tau=5$ & $\tau=10$\\
    \midrule
     \textsc{\textsc{EGNN}}& 19.99 & 21.37 & 8.21 & 1\% & 1\% & 3\% \\
     \textsc{\bf{\textsc{SBAlign}}} (10, 10) & 3.80 & 4.98 & 3.95 & 0\% & 69\% & 93\%\\
     \textsc{\bf{\textsc{SBAlign}}} (10, 100) & 3.81 & 5.02 & 3.96 & 0\% & 70\% & 93\%\\
     \bottomrule \vspace{-15pt}
    \end{tabular}
}
\end{table}

\para{Future outlook} In this section, we presented a proof of concept application of \textsc{SBAlign} for modelling conformational changes in proteins during docking. associated with the protein docking task. A combination of \textsc{SBAlign} for conformational change modeling, with more recent methods for rigid-protein docking \citep{ketata2023diffdock} can provide a complete solution for the protein docking task, which we leave to future work.

\vspace{-5pt}
\section{Conclusion}
\label{sec:conclusion}
\vspace{-5pt}
In this paper, we propose a new framework to tackle the interpolation task with aligned data via \aclp{DSB}. Our central contribution is a novel algorithmic framework derived from the Schr\"odinger bridge theory and Doob's $h$-transform. Via a combination of the two notions, we derive novel loss functions which, unlike all prior methods for solving \aclp{DSB}, do not rely on the \acl{IPF} procedure and are hence numerically stable. We verify our proposed algorithm on various synthetic and real-world tasks and demonstrate noticeable improvement over the previous state-of-the-art, thereby substantiating the claim that data alignment is a highly relevant feature that warrants further research.

\section*{Acknowledgements}
This publication was supported by the NCCR Catalysis (grant number 180544), a National Centre of Competence in Research funded by the Swiss National Science Foundation as well as the European Union’s Horizon 2020 research and innovation programme 826121. We thank Caroline Uhler for introducing us to the dataset by \citet{weinreb2020lineage}, which was instrumental in this research.

\bibliography{references}

\clearpage
\numberwithin{equation}{section}		%
\numberwithin{lemma}{section}		%
\numberwithin{proposition}{section}		%
\numberwithin{theorem}{section}		%
\onecolumn %
\title{Aligned Diffusion Schr\"odinger Bridges \\ (Supplementary Material)}
\maketitle

\appendix

\footnotetext[1]{Equal contribution.}
\section{Additional Results} 
\label{app:more_results}

\subsection{Variance Reduction}
\label{app:sec:var_reduction}

In this paragraph, we elaborate on the need to parametrize also Doob's $\doob$ function, along with the drift $\fdrift$.
Introducing $m^{\phi}$ removes the need to evaluate \eqref{eq:softdoob} which is difficult to approximate in practice on high-dimensional spaces. This equation amounts, in fact, to a Gaussian Kernel Density Estimation of the conditional probability $\prob(X_1 = \x_1\vert X_t = \x)$ along (unconditional) paths obtained from \eqref{eq:SB-SDE}.
Faithful approximations of \eqref{eq:softdoob} would, therefore, require:
\begin{itemize}
    \item good-quality paths, which are scarce at the beginning of training when the drift $\fdrift^\theta$ has not yet been learned;
    \item exponentially many trajectories (in the dimension of the state space);
    \item that points $x_1$ (obtained from conditional trajectories, Eq. \ref{eq:SB-SD-conditioned}) be reasonably close to $x_1$ (obtained from unconditional trajectories, Eq. \ref{eq:SB-SDE});
\end{itemize}
Even if all the above conditions were satisfied, the quantity $\doob(x) = \prob(X_1 = \x_1\vert X_t = \x)$ would still be challenging to directly manipulate. 
It is, in fact, much smaller at earlier times $t$ (see Table \ref{tab:doob_magnitude_evo}), since knowledge of the far past has a weaker influence on the location $X_1$ of particles at time $t=1$. 
Precision errors at $t \approx 0$ would then be amplified when computing the score of $\doob$ (i.e, $\nabla \log \doob$) --which appears in the loss \eqref{eq:loss_modified}-- and accumulate over timesteps, eventually leading trajectories astray.
By directly parameterizing the score, we instead sidestep this problem. The magnitude of $m^\phi_t \approx \nabla \log \doob$ can, in fact, be more easily controlled and regularized.
\begin{table}[h]
    \centering
    \begin{tabular}{l|ccccccc}
    \toprule
          & \multicolumn{7}{c}{\textbf{Time $t$}} \\
          & \textbf{0} & \textbf{0.15} & \textbf{0.30} & \textbf{0.45} & \textbf{0.60} & \textbf{0.75} & \textbf{0.90} \\
    \midrule
         \textbf{Mean $\doobs$ value} &  2.92e-14	& 4.03e-13 & 2.54e-11 & 1.72e-09 & 1.47e-07 & 2.66e-05 & 8.53e-3 \\
    \bottomrule
    \end{tabular}
    \caption{Average $\doobs$ values along paths, at different timesteps. $\prob(X_1 = \x_1\vert X_t = \x)$ ranges over 11 orders of magnitude across the time interval and is smallest when $t \approx 0$.}
    \label{tab:doob_magnitude_evo}
\end{table}

\section{Datasets} 
\label{app:datasets}
\subsection{Synthetic Datasets}
In the following, we provide further insights and experimental results in order to access the performance of \textsc{SBalign} in comparison with different baselines and across tasks of various nature.
For each dataset, we describe in detail its origin as well as preprocessing and featurization steps.

\begin{figure}[ht]
    \centering
    \includegraphics[width=0.45\textwidth]{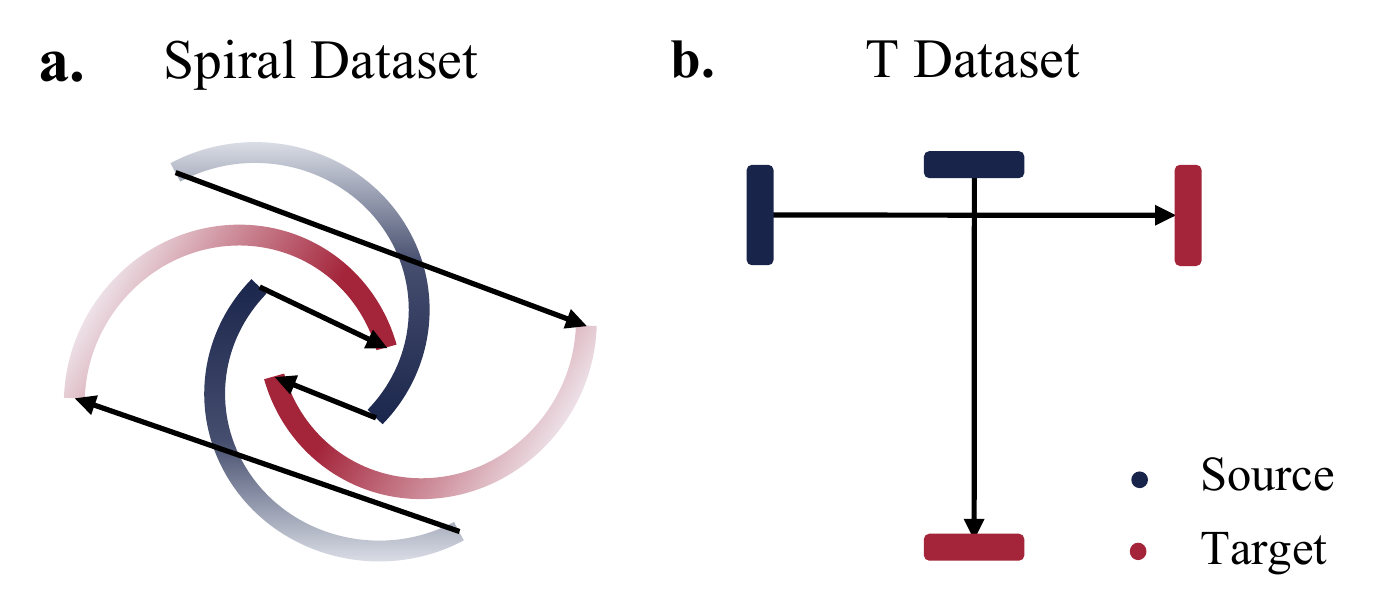}
    \caption{Initial (\textit{blue}) and final (\textit{red}) marginals for the two toy datasets \textbf{(a)} moon and \textbf{(b)} T, together with arrows indicating a few alignments}
    \label{fig:synthetic_datasets}
\end{figure}

\para{Moon dataset}
The \texttt{moon} toy dataset (Fig.~\ref{fig:synthetic_datasets}a) is generated by first sampling $\distend$ and then applying a clockwise rotation of $233^\circ$ around the origin to obtain $\distinit$.
The points on the two semi-circumferences supporting $\distend$ are initially placed equally-spaced along each semi-circumference and then moved by applying additive Gaussian noise to both coordinates. While classic generative models will choose the shortest path and connect ends of both moons closest in Euclidean distance, only methods equipped with additional knowledge or insight on the intended alignment will be able to solve this task.

\para{T dataset}
This toy dataset (Fig.~\ref{fig:synthetic_datasets}b) is generated by placing an equal amount of samples at each of the four extremes of a T-shaped area having ratio between \textit{x} and \textit{y} dimensions equal to 51/55.
If run with a Brownian prior, classical \acp{SB} also fail on this dataset because they produce swapped pairings: i.e., they match the left (\textit{resp.} top) point cloud with the bottom (\textit{resp.} right) one.
At the same time, though, this dataset prevents reference drifts with simple analytical forms (such as spatially-symmetric or time-constant functions) from fixing classical \acp{SB} runs. It therefore illustrates the need for general, plug-and-play methods capable of generating approximate reference drifts to use in the computation of classical \acp{SB}.

\subsection{Cell Differentiation Datasets}
\label{app:datasets-cell_differentiation}
\begin{figure}[ht]
    \centering
    \includegraphics[width=0.65\linewidth]{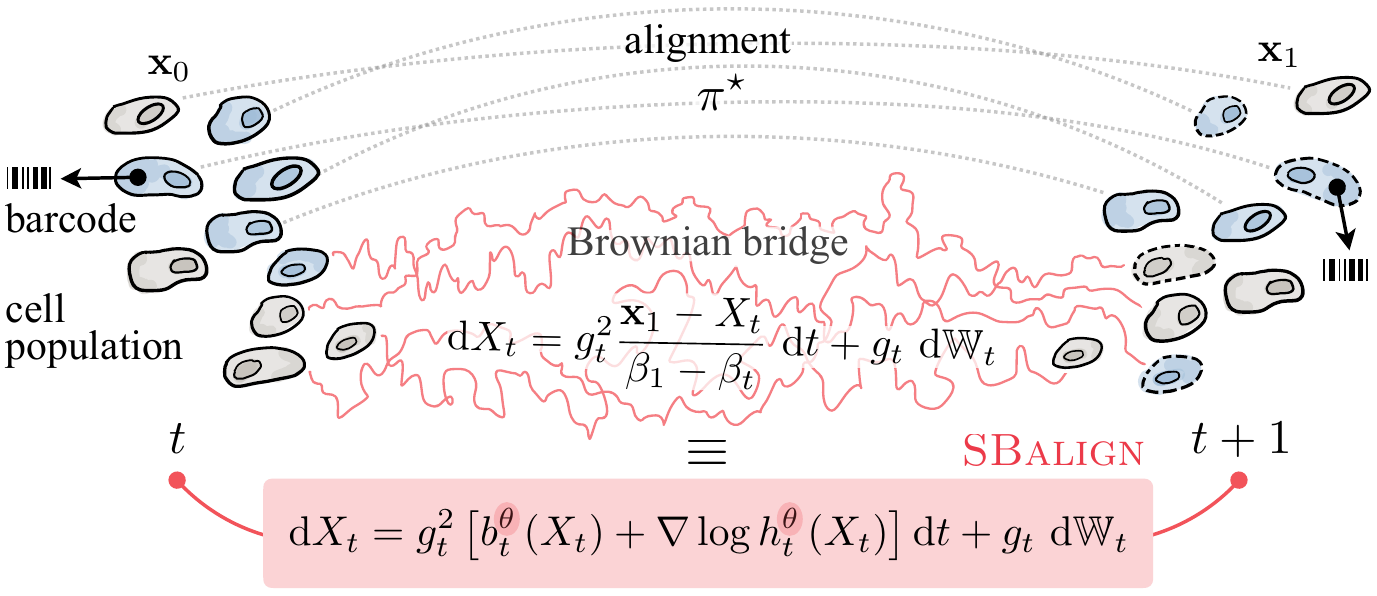}
    \caption{Overview of \textsc{SBalign} in the setting of cell differentiation with the goal of learning the evolutionary process that morphs a population from its stat at $t$ to $t+1$. Through genetic tagging (i.e., barcodes) we are able to trace progenitor cells at time point $t$ into their descendants $t+1$. This provides us with an alignment between populations at consecutive time steps. Our goal is then to recover a stochastic trajectory from $\x_0$ to $\x_1$. To achieve this, we connect the characterization of a SDE conditioned on $\x_0$ and $\x_1$ (utilizing the Doob's \emph{$h$-transform}) with that of a Brownian bridge between $\x_0$ and $\x_1$ (classical Schr\"odinger bridge theory), leading to a simpler training procedure with lower variance and strong empirical results.}
    \label{fig:overview_cells}
\end{figure}

\para{Dataset description}
We obtain the datapoints used in our cell differentiation task from the dataset generated by \cite{weinreb2020lineage}, which contains 130861 observations/cells.
We follow the preprocessing steps in \citet{bunne2021learning} and use the Python package \texttt{scanpy} \citep{wolf2018scanpy}. After processing, each observation records the level of expression of 1622 different highly-variable genes as well as the following meta information per cell:
\begin{itemize}
    \item a \texttt{timestamp}, expressed in days and taking values in \{2, 4, 6\};
    \item a \texttt{barcode}, which is a short DNA sequence that allows tracing the identity of cells and their lineage by means of single-cell sequencing readouts;
    \item an additional \texttt{annotation}, which describes the current differentiation fate of the cell.
\end{itemize}
\begin{figure}[ht]
    \centering
    \includegraphics[width=.7\textwidth]{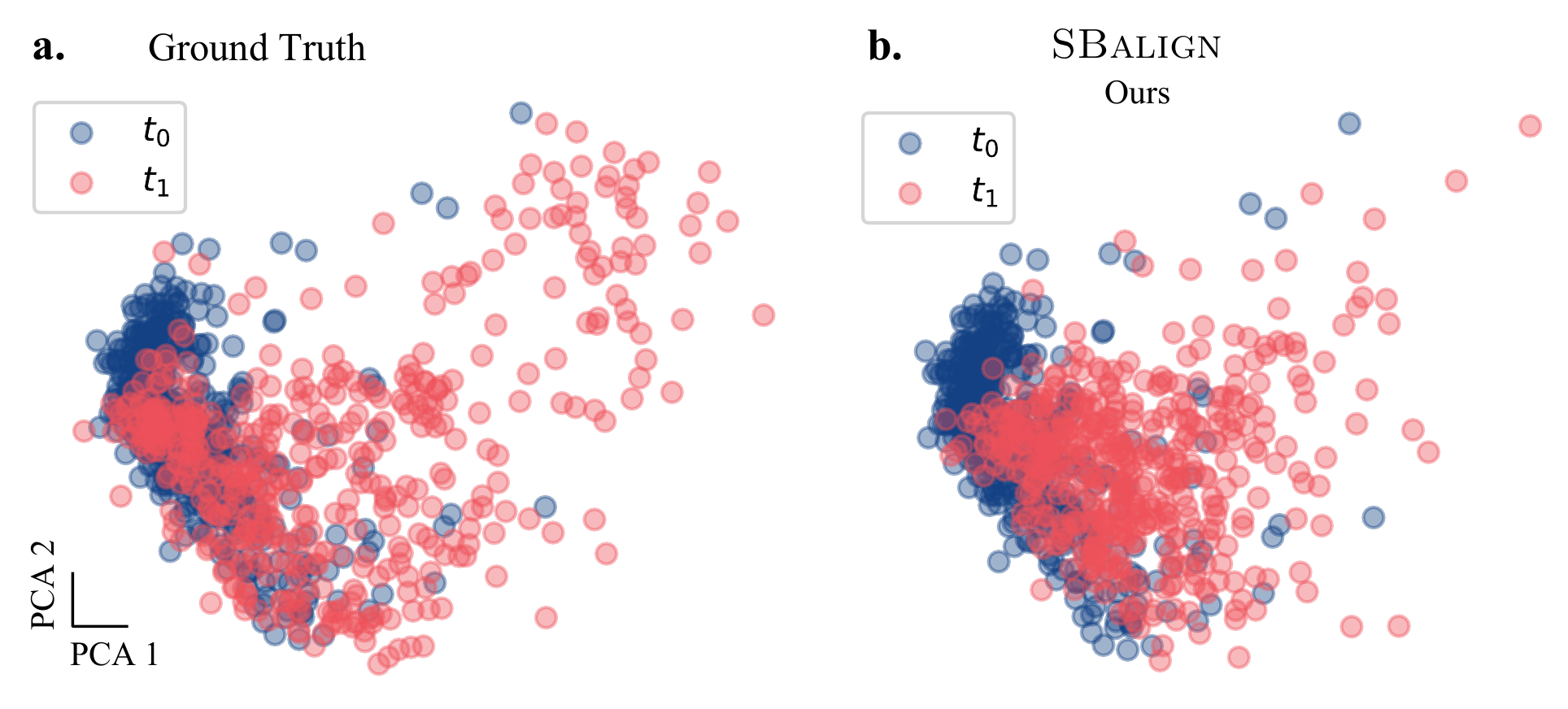}
    \caption{Distribution of the cell population (i.e., marginals) at time $t=t_0$ and $t=t_1$ for (\textbf{a}) the ground truth, and (\textbf{b}) \textsc{SBalign}, after projection along their first two principal components.}
    \label{fig:results_cell_marginals}
\end{figure}

\para{Dataset preparation}
We only retain cells with barcodes that appear both on days 2 and 4, taking care of excluding cells that are already differentiated on day 2.
We construct matchings by pairing cells measured at two different times but which share the barcode. 
Additionally, we filter cells to make sure that no one appears in more than one pair.
To reduce the very high dimensionality of these datapoints, we perform a PCA projection down to 50 components.

We end up with a total of 4702 pairs of cells, which we partition into train, validation, and test sets according to the split 80\%/10\%/10\%.

\subsection{Conformational Changes in Protein Docking}
\label{app:datasets-protein_conf}

\para{Dataset description} For the task of predicting protein conformational changes, we utilize the D3PM dataset. The dataset consists of both unbound and bound structures for 4330 proteins, under different types of protein motions. The PDB IDs were downloaded from \url{https://www.d3pharma.com/D3PM/}. For the PDB IDs making up the dataset, we download the corresponding (.cif) files from the Protein Data Bank. 

\para{Dataset preparation} For the scope of this work, we only focus on protein structure pairs, where the provided RMSD between the C$\alpha$ carbon atoms is $>3$\r{A}, amounting to 2370 examples in the D3PM dataset. For each pair of structures, we first identify common residues, and compute the RMSD between C$\alpha$ carbon atoms of the common residues after superimposing them using the Kabsch \citep{kabsch1976solution} algorithm, and only accept the structure if the computed C$\alpha$ RMSD is within a certain margin of the provided C$\alpha$ RMSD. The rationale behind this step was to only retain examples where we could reconstruct the RMSD values provided with the dataset. Common residues are identified through a combination of residue position and name. This step is however prone to experimental errors, and we leave it to future steps to improve the common residue identification step (using potentially, a combination of common subsequences and/or residue positions). 

After applying the above preprocessing steps, we obtain a dataset with 1591 examples, which is then split into a train/valid/test split of 1291/150/150 examples respectively. The structures used in training and inference are the Kabsch superimposed versions, therefore ensuring that the Brownian bridges are sampled between the unbound and bound states of the proteins, and no artifacts are introduced by 3D rotations and translations, which do not contribute to conformational changes.

\para{Featurization} Following standard practice and for memory and computational efficiency, we only use the C$\alpha$ coordinates of the residues to represent our protein structures instead of the full-atom structures. For each amino acid residue, we compute the following features: a one hot encoding of the amino acid identity $f_e$ of size $23$, hydrophobicity $f_h \in [-4.5, 4.5]$, volume $f_v \in [60.1, 227.8]$, the charge $f_c \in \{-1, 0, 1\}$, polarity $f_p \in \{0, 1\}$, and whether the amino acid residue is a hydrogen bond donor $f_d \in \{0, 1\}$ or acceptor $f_a \in \{0, 1\}$. The hydropathy and volume features are expanded into a radial basis with interval sizes $0.1$ and $10$ respectively. To equip the model with a notion of time, we use a sinusoidal embedding of time $\phi(t)$ of embedding dimensionality $32$. These are concatenated to the amino acid features to form our input features for the amino acid residues. The edge features consist of a radial basis expansion of the distances between the residues. We also compute the spherical harmonics of the edge vectors between the residues, which is used in the tensor product message passing layers.

\para{Position at t} For any time $t$, we sample the positions of the C$\alpha$ atoms using the Brownian Bridge - given the coordinates $\x_0$ at $t=0$ and the coordinates $\x_1$ at $t=1$ with a Brownian bridge between $\x_0$ and $\x_1$, we know that $x_t \sim \mathcal{N}\left(x_t; (1-t)\x_0 + t\x_1, t(1-t)\right)$.

\section{Experimental Details}

In the following, we provide further experimental details on the chosen evaluation metrics, network architectures, and hyperparameters.

\subsection{Evaluation Metrics} 
\label{app:metrics}

\subsubsection{Cell Differentiation}
\label{app:metrics-cell_differentation}
For fairness of comparison between our method and the baseline (\textsc{fbSB}) ---which only works at the level of distribution of cells--- we also consider three evaluation metrics (i.e., $\We$, MMD and $\ell_2$) that capture the similarity between the end marginal $\distend$ and our prediction $\pi^\star_1$, irrespective of matchings.

In what follows, we denote with $\hat{\nu}$ the predicted end marginal $\pi^\star_1$ ---i.e., the predicted status of cells at day 4--- and with $\nu$ the distribution of observed transcriptomes.

\para{Wasserstein-2 distance} We measure accuracy of the predicted target population $\hat{\nu}$ to the observed target population $\nu$ using the entropy-regularized Wasserstein distance \citep{cuturi2013sinkhorn} provided in the \texttt{OTT} library \citep{jax2018github,cuturi2022optimal} defined as
\begin{equation}\label{eq:reg-ot}
\end{equation}
where $H(\bP) \defeq -\sum_{ij} \bP_{ij} (\log \bP_{ij} - 1)$ and the polytope $U(\hat{\nu},\nu)$ is the set of $n\times m$ matrices $\{\bP\in\mathbb{R}^{n \times m}_+, \bP\mathbf{1}_m = \hat{\nu}, \bP^\top\mathbf{1}_n=\nu\}$.

\para{Maximum mean discrepancy} Kernel maximum mean discrepancy~\citep{gretton2012kernel} is another metric to measure distances between distributions, i.e., in our case between predicted population $\hat{\nu}$ and observed one $\nu$.
Given two random variables $x$ and $y$ with distributions $\hat{\nu}$ and $\nu$, and a kernel function $\omega$, \citet{gretton2012kernel} define the squared MMD as:
\begin{equation*}
    \text{MMD}(\hat{\nu},\nu; \omega) = \mathbb{E}_{x,x^\prime}[\omega(x, x^\prime)] + \mathbb{E}_{y,y^\prime}[\omega(y, y^\prime)] - 2\mathbb{E}_{x,y}[\omega(x, y)].
\end{equation*}
We report an unbiased estimate of $\text{MMD}(\hat{\nu},\nu)$, in which the expectations are evaluated by averages over the population particles in each set. We utilize the RBF kernel, and as is usually done, report the MMD as an average over the length scales: $2, 1, 0.5, 0.1, 0.01, 0.005$.

\para{Perturbation signature $\ell_2$}
A common method to quantify the effect of a perturbation on a population is to compute its perturbation signature \citep[(PS)]{stathias2018drug}, computed via the difference in means between the distribution of perturbed states and control states of each feature, e.g., here individual genes. $\ell_2$(PS) then refers to the $\ell_2$-distance between the perturbation signatures computed on the observed and predicted distributions, $\nu$ and $\hat{\nu}$. The $\ell_2$(PS) is defined as
\begin{equation*}
    \text{PS}(\nu, \mu) = \frac{1}{m}\sum_{y_i \in \nu}{y_i} - \frac{1}{n}\sum_{x_i \in \mu}{x_i},
\end{equation*}
where $n$ is the size of the unperturbed and $m$ of the perturbed population.
We report the $\ell_2$ distance between the observed signature $\text{PS}(\nu, \mu)$ and the predicted signature $\text{PS}(\hat{\nu}, \mu)$, which is equivalent to simply computing the difference in the means between the observed and predicted distributions.

\para{RMSD}
To measure the quality of matchings sampled from \textsc{SBalign} $(\hat{x}^i_0, \hat{x}^i_1)$ ---compared to the observed ones $(x^i_0, x^i_1)$--- we compute:
\begin{equation}
    \text{RMSD}(\{x^i_1\}^n,\{\hat{x}^i_1\}^n) = \sqrt{\frac{1}{n}\sum^n_{i=1} \lVert x^i_1 - \hat{x}^i_1\rVert^2}
\end{equation}
which, when squared, represents the mean of the square norm of the differences between predicted and observed statuses of the cells on day 4.

\para{Cell type classification accuracy}
We assess the quality of \textsc{SBalign} trajectories by trying to predict the differentiation fate of cells, starting from (our compressed representation of) their transcriptome.
For this, we train a simple MLP-based classifier on observed cells and use it on the last time-frame of trajectories sampled from \textsc{SBalign} to infer the differentiation of cells on day 4.
We use the classifier \texttt{MLPClassifier} offered by the library \texttt{scikit-learn} with the following parameters:
\begin{itemize}
    \item 2 hidden layers, each with a hidden dimension of 50,
    \item the \textit{}{logistic} function as non-linearity
    \item $\ell_2$ norm, regularization with coefficient $0.1$.
\end{itemize}

We report the subset accuracy of the predictions on the \textit{test} set, measured as the number of labels (i.e., cell types) coinciding with the ground truth.

\subsection{Network Architectures}
\label{app:architectures}

\subsubsection{Cell Differentiation and Synthetic Datasets}
\label{app:architectures-cell_diff_synthetic}

We parameterize both $b^\theta(t, X_t)$ and $m^\phi(t, b_t, X_t)$ using a model composed of:
\begin{enumerate}
    \item \textbf{\texttt{x\_enc}}: 3-layer MLP performing the expansion of spatial coordinates (or drift) into hidden states (of dimension 64 to 256);
    \item \textbf{\texttt{t\_enc}}: sinusoidal embedding of time (on 64 to 256 dimensions), followed by a two layer MLP;
    \item \textbf{\texttt{mlp}}: 3-layer MLP which maps the concatenation of embedded spatial and temporal information (output of modules 1 and 2 above) to drift magnitude values along each dimension.
\end{enumerate}

After every linear layer (except the last one), we apply a non-linearity and dropout (level 0.1).
In all the experiments, we set the diffusivity function $g(t)$ in \eqref{eq:SB-SDE} to a constant $g$, which is optimized (see \S~\ref{app:hyperparams}).

\subsubsection{Conformational Changes in Protein Docking}
\label{app:architectures-protein_conf}

As our architecture $b^{\theta}_t(X_t)$ suitable for approximating the true drift $b_t$, we construct a graph neural networks with tensor-product message passing layers using \texttt{e3nn} \citep{thomas2018tensor, geiger2022e3nn}. To build the graph, we consider a maximum of 40 neighbors --located within a radius of 40\r{A} for each residue. The model is SE(3) equivariant and receives node and edge features capturing relevant residue properties, and distance embeddings.

\subsection{Hyperparameters}
\label{app:hyperparams}

In the following, we will provide an overview of the selected hyperparameters as well as chosen training procedures.

\subsubsection{Synthetic Tasks}
\label{app:hyperparams-synthetic}

We perform hyper-parameter optimization using the Python package \texttt{ray.tune} \citep{liaw2018tune} on:
\begin{itemize}
    \item \textbf{activation}, chosen among \texttt{leaky\_relu}, \texttt{relu}, \texttt{selu} and \texttt{silu} as implemented in the Python library \texttt{PyTorch} \citep{pytorch2019paszke}. We find \texttt{selu} to achieve marginally better performance on toy datasets.
    \item \textbf{g}, the value of the diffusivity constant, chosen among $\{1, 2, 5, 10\}$. We find $g=1$ to yield optimal results.
\end{itemize}

\subsubsection{Cell Differentiation}
\label{app:hyperparams-cell_differentiation}

We perform hyper-parameter optimization using the Python package \texttt{ray.tune} \citep{liaw2018tune} on:
\begin{itemize}
    \item \textbf{activation}, chosen among \texttt{leaky\_relu}, \texttt{relu}, \texttt{selu}, and \texttt{silu} as implemented in the Python library \texttt{PyTorch} \citep{pytorch2019paszke}. We observe that \texttt{silu} brings noticeable performance improvements on the cell differentiation dataset.
    \item \textbf{g}, the value of the diffusivity constant, chosen among $\{0.01, 0.1, 0.8, 1, 1.2, 2, 5\}$. We find $g=1$ to yield optimal results.
\end{itemize}

\subsubsection{Conformational Changes in Protein Docking}
\label{app:hyperparams-protein_conf}

We use \texttt{AdamW} as our optimizer with a initial learning rate of $0.001$, and training batch size of $2$. For each protein pair, we sample $10$ timepoints in every epoch, so the model sees realizations from different timepoints of the corresponding Brownian Bridge. This was done to improve the training speed. We use a regularization strength of $1.0$ for $m^{\phi}$ for all $t$. Inference on the validation set using training is carried out using the exponential moving average of parameters, and the moving average is updated every optimization step with a decay rate of $0.9$. The model training is set to a maximum of $1000$ epochs but training is typically stopped after $200$ epochs beyond which no improvements in the validation metrics are observed. 

Our model has 0.54M parameters and is trained for 200 epochs. After every epoch, we simulate trajectories on the validation set using our model and compute the mean RMSD. The best model selected using this procedure is used for inference on the test set. The baseline \textsc{EGNN} model has 0.76M parameters and is trained for 1000 epochs.

\section{Reproducibility}
Code utilized in this publication can be found at \url{https://github.com/vsomnath/aligned_diffusion_bridges}, with a mirror at \url{https://github.com/IBM/aligned_diffusion_bridges}.

\end{document}